\documentclass{article}
\usepackage{spconf,amsmath,graphicx}
\usepackage{multirow}

\title{Image Forgery Localization based on \\ Multi-Scale Convolutional Neural Networks}
\name{Yaqi~Liu,~Qingxiao~Guan,~Xianfeng~Zhao,~and~Yun~Cao\thanks{This work was supported by the NSFC under U1536105 and U1636102, and National Key Technology R\&D Program under 2014BAH41B01, 2016YFB0801003 and 2016QY15Z2500.}}
\address{1. State Key Laboratory of Information Security, Institute of Information Engineering, \\
Chinese Academy of Sciences, Beijing 100093, China \\
2. School of Cyber Security, University of Chinese Academy of Sciences, Beijing 100093, China}
\begin{document}
%
\maketitle
\begin{abstract}
In this paper, we propose to utilize Convolutional Neural Networks (CNNs) and the segmentation-based multi-scale analysis to locate tampered areas in digital images. First, to deal with color input sliding windows of different scales, a unified CNN architecture is designed. Then, we elaborately design the training procedures of CNNs on sampled training patches. With a set of robust multi-scale tampering detectors based on CNNs, complementary tampering possibility maps can be generated. Last but not least, a segmentation-based method is proposed to fuse the maps and generate the final decision map. By exploiting the benefits of both the small-scale and large-scale analyses, the segmentation-based multi-scale analysis can lead to a performance leap in forgery localization of CNNs. Numerous experiments are conducted to demonstrate the effectiveness and efficiency of our method.
\end{abstract}
\section{Introduction}
\label{sec:Intro}
%
%
%
%

\begin{figure}[t]
  \centering
  \centerline{\includegraphics[width=8cm]{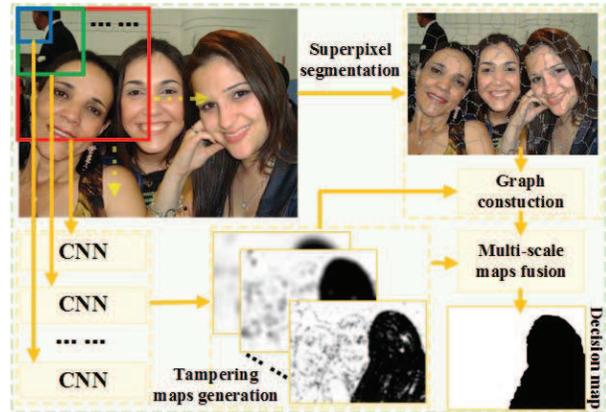}}
\caption{\ The framework of MSCNNs. Note that the sliding windows (blue, green, red squares) and superpixels do not indicate their real sizes.
}\label{fig:framework}
\end{figure}

Image forgery localization is one of the most challenging tasks in digital image forensics \cite{li2017image}. Different from forgery detection which simply discriminates whether a given image is pristine or fake, image forgery localization attempts to detect the accurate tampered areas \cite{korus2016multi}. Since forgery localization needs to conduct pixel-level analyses, it is more difficult than the conventional forgery detection task.

Different clues are investigated to locate the tampered areas, e.g., the photo-response nonuniformity noise (PRNU) \cite{chen2008determining}, the artifacts of color filter array \cite{ferrara2012image}, the traces left by JPEG coding \cite{bianchi2012image}, the near-duplicate image analysis \cite{gaborini2014multi}, and copy-move forgery detection \cite{cozzolino2015efficient}, etc. The tampering operations inevitably distort some inherent relationships among the adjacent pixels, features motivated by steganalysis \cite{fridrich2012rich} are frequently adopted to localize tampered areas \cite{cozzolino2016single,li2017image}. In 2013, IEEE Information Forensics and Security Technical Committee (IFS-TC) established the First IFS-TC Image Forensics Challenge \cite{IFSTC}. In the second phase, a complicated and practical situation for evaluating the performance of forgery localization was set up. The winner \cite{cozzolino2014image} and successors \cite{gaborini2014multi,li2017image} combined different clues to achieve high scores. As far as we know, the former best F1-score using a single clue from statistical features was achieved in \cite{li2017image} which was based on color rich models \cite{goljan2014rich} and the ensemble classifier \cite{kodovsky2012ensemble} (SCRM+LDA).

We focus on forgery localization utilizing statistical features extracted by Convolutional Neural Networks (CNNs) \cite{krizhevsky2012imagenet}. Booming in computer vision tasks, CNNs are also applied in image forensics. In \cite{chen2015median}, CNNs are applied in median filtering image forensics. In \cite{bayar2016deep}, a novel constrained convolutional layer is utilized to suppress the content of the image, and CNNs are adopted to detect multiple manipulations. In \cite{rao2016deep}, a CNN with SRM kernels \cite{fridrich2012rich} for the first layer initialization is adopted for forgery detection. In \cite{cozzolino2017recasting}, they show that residual-based descriptors can be regarded as a simple constrained CNN which can conduct forgery detection and localization. Numerous meaningful works have been done to improve the performance of image forensics by adopting CNNs, they try to construct different CNN architectures. While in computer vision tasks, typical CNNs, e.g. AlexNet \cite{krizhevsky2012imagenet}, VGG \cite{simonyan2014very}, ResNet \cite{he2016identity}, etc., are directly adopted for different purposes \cite{liu2017listnet}, and they mainly focus on the preprocessing and postprocessing. This kind of adoptions accelerate the development of many computer vision tasks. Thus, instead of designing a totally novel CNN, we adopt and modulate the state-of-the-art CNNs \cite{xu2016structural} to construct our framework for image forgery localization. More powerful CNNs can also be adopted in the proposed framework in the future.

In this paper, an image forgery localization method based on Multi-Scale Convolutional Neural Networks (MSCNNs) is proposed, as shown in Figure \ref{fig:framework}. In our method, sliding windows of different scales are put into a set of CNNs to generate real-valued tampering possibility maps. Then, based on the graph constructed on superpixels \cite{achanta2012slic}, we can generate the final decision map by fusing those possibility maps. The contributions are two-fold: First, we propose to utilize multi-scale CNNs to detect forged regions. A unified CNN architecture is formulated for color patches, and multi-scale CNNs are treated as a set of ``weak" classifiers to fully exploit the benefits of both the small-scale and large-scale analyses. Second, based on the fusion method in \cite{korus2016multi}, the segmentation-based fusion method is proposed to efficiently process images of different sizes. Maps fusion based on conditional random fields is conducted on the superpixel-level graph, and two strategies for superpixel-level tampering possibility maps generation are proposed and compared.

On the IFS-TC dataset, MSCNNs can achieve the best performance among the forgery localization methods which merely utilize one kind of clue for splicing detection. To the best of our knowledge, only three methods, i.e., the winner \cite{cozzolino2014image} and successors \cite{gaborini2014multi,li2017image}, can achieve higher scores than MSCNNs, but they all combine multiple different clues, e.g. statistical features, copy-move clues etc. The proposed MSCNNs only utilizes statistical features extracted by CNNs and can be further improved by adopting other clues. Besides, to demonstrate the robustness of the proposed framework, we also conduct experiments on another dataset, i.e. Realistic Tampering Dataset (RTD) \cite{korus2016multi,Korus2016WIFS}.

The rest of the paper is structured as follows. In Section \ref{sec:method}, we elaborate the proposed method. In Section \ref{sec:Experiments}, experiments are conducted. In Section \ref{sec:DiscussionConclusion}, we draw conclusions.

\section{Method}
\label{sec:method}

\subsection{CNNs architecture}
\label{sec:CNNs}

Our motivation is that we want to replace the SCRM+LDA \cite{li2017image} with the end-to-end CNNs to estimate the tampering probability of a given patch. Adopting the sliding window manner, we can give the tampering possibility map of the investigated image. The CNNs proposed in \cite{xu2016structural} achieve the state-of-the-art performance for steganalysis on gray-scale images. Considering the close relationship between image forensics and steganalysis, we adopt this kind of CNNs as the basic architecture in our work.  In the first layer of their CNNs, a single high pass filter (we call it the base filter) is utilized to suppress the image content. In our work, to deal with color patches, two kinds of base filters are tested:

(1) Fixed SRM kernels: the base filters are fixed, and set as the SRM kernels \cite{fridrich2012rich}. In \cite{rao2016deep}, $30$ SRM kernels are adopted for the initialization of the first layer of their CNNs. We adopt all the SRM kernels as fixed base filters, and leave the task of validating their effectiveness to the backend network. Referring to \cite{rao2016deep}, the $30$ SRM kernels are formulated as $5\times 5$ matrixes $\{\mathbf{F}_1,\cdots \mathbf{F}_{30}\}$ with zero-valued unused elements. The inputs are three-channel color patches, so we need $30\times 3$ filters to generate $30$ feature maps. For the $j$th feature map ($j\in \{ 1,2,\cdots 30 \}$), the corresponding filters are set as $\{\mathbf{F}_1^{j},\mathbf{F}_2^{j},\mathbf{F}_3^{j}\}=\{\mathbf{F}_{3k-2},\mathbf{F}_{3k-1},\mathbf{F}_{3k}\}$, where $k=((j-1)~{\rm mod}~10)+1$.

(2) Constrained filters: in \cite{bayar2016deep}, a kind of constrained filter was proposed for manipulation detection. Here, we adopt it for forgery localization. The constraint means that the filter weight at the center $f(0,0)=-1$, and $\sum_{r,c\ne 0}f(r,c)=1$, $f(r,c)$ denotes the element in the base filter $\mathbf{F}$. For fair comparisons, $90$ $5\times5$ constrained filters are adopted.

\begin{figure*}[htp]
  \centering
  \centerline{\includegraphics[width=15.8cm]{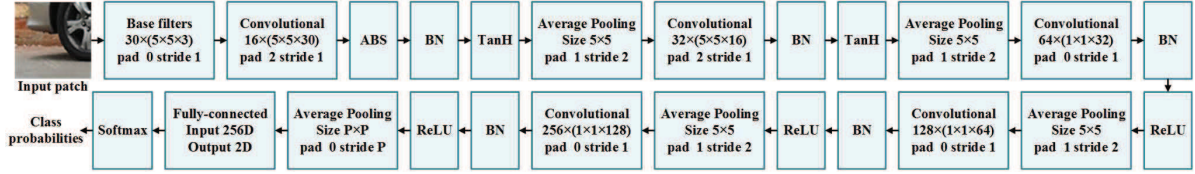}}
\caption{\ The architecture and parameters of the unified CNNs.
}\label{fig:CNNs}
\end{figure*}

As we adopt $90$ base filters, we modulate the parameters of CNNs in \cite{xu2016structural}, and the unified CNN architecture can be depicted as Figure \ref{fig:CNNs}. For different scales of input patches, we only need to change $P$ in the last average pooling layer, ensuring that the input of the fully-connected layer is a $256$-dimensional vector. Based on the CNN depicted in Figure \ref{fig:CNNs}, we can train a set of CNN detectors with input patches of different scales. The detailed training procedures are introduced in Section \ref{sec:Experiments}.

\subsection{Maps generation}
\label{sec:MG}

For each input image, it is analysed by the sliding window of the scale as $s\times s$ with a stride of $st$ based on the CNN detectors described in Section \ref{sec:CNNs}. Then, we can get the tampering possibility map $\hat{\mathbf{M}}_{s}$ of size $h_{s}\times w_{s}$, where $h_{s}=\lfloor (h-s)/st \rfloor+1$ and $w_{s}=\lfloor (w-s)/st \rfloor+1$, $h$ and $w$ denote the height and width of the input image, and $\lfloor \cdot \rfloor$ denotes the floor function. The elements in $\hat{\mathbf{M}}_{s}$ denote the probabilities of the corresponding patches being fake. In order to get the possibility map $\mathbf{M}_{s}$ with the same size as the input image, the element $m_{i,j}^{s}$ in $\mathbf{M}_{s}$ is computed as:
\begin {equation}\label{eq:feamatri}
m_{i,j}^{s}=\frac{1}{K}\sum\nolimits_{k=1}^{K}\hat{m}_{k}^{s}
\end {equation}
where $K$ is the number of patches containing pixel $I_{i,j}$, and $\hat{m}_{k}^{s}$ denotes the corresponding value in $\hat{\mathbf{M}}_{s}$. Inevitably, for some pixels, $K$ is equal to $0$, and the pixels always appear around the edges of the image. We simply set the same probabilities as the nearest pixels whose $K\ne0$. Since we have a large stride $st$, there are mosaic artifacts in the possibility map generated by formula (\ref{eq:feamatri}). Naturally, it is expected that the map for an image tends to be smoother \cite{li2017image}. To smooth the possibility map, the mean filtering is applied as:
\begin {equation}\label{eq:smfeamatri}
\bar{m}_{i,j}^{s}=\frac{1}{s\times s}\sum_{i'=-\frac{s}{2}}^{\frac{s}{2}-1}\sum_{j'=-\frac{s}{2}}^{\frac{s}{2}-1}m_{i+i',j+j'}^{s}
\end {equation}
where $s$ is the size of corresponding patches. Thus, we can get the smoothed possibility map $\bar{\mathbf{M}}_{s}$ with elements as $\bar{m}_{i,j}^{s}$.

\subsection{Maps fusion}
\label{sec:MF}

With the analyses of multi-scale CNNs detectors, we can get a set of tampering possibility maps $\{\bar{\mathbf{M}}_{s}\}$ for each image, and $s$ denotes the scales of input patches. The final task is to fuse possibility maps to exploit the benefits of multi-scale analyses. In \cite{korus2016multi}, the multi-scale analysis in PRNU-based tampering localization was proposed. By minimizing an energy function, possibility maps fusion is formulated as a random-field problem where decision fusion resolves to finding an optimal labeling of authentication units. The optimization problem is solved by the graph cut algorithm whose worst case running time complexity is $O(ev^2)$ \cite{boykov2004experimental}, where $v$ is the number of nodes in the graph and $e$ is the edge number. They consider a 2nd-order neighborhood, which means that $e\approx4v$, so the complexity of the method is $O(v^3)$. They adopt pixels as the nodes in the graph, thus the computing time of the large image is almost unacceptable. So we propose to construct graphs on superpixels, and find the optimal labels on the superpixel level.

Simple linear iterative clustering (SLIC) \cite{achanta2012slic} is a commonly used efficient superpixel segmentation method, and we adopt SLIC to conduct oversegmentation on the investigated color images. The complexity of SLIC is linear, i.e. $O(v)$, and it is easy to generate superpixels by SLIC for large images. In the computer vision tasks, images are usually segmented into hundreds of superpixels. In the task of tampering possibility maps fusion, large superpixels can lead to information loss. Thus, thousands of superpixels must be generated in our task. Then, a graph on the superpixels is constructed, each superpixel is treated as a node in the graph and the adjacent superpixels are connected by an edge. The number of graph nodes is around several thousand, which is much easier to compute by the graph cut algorithm. Besides the efficiency of the superpixel-level computation, the segmentation-based method can also well adhere to the real boundaries, and avoid mislabeling of homogeneous pixels, resulting in the performance improvement.

As for the superpixel-level tampering possibility maps $\mathbf{M}_{s}^{sup}$ generation, two strategies are proposed and compared. The one is ``mean", and the tampering possibility $m_{sup_l}^{s}$ of superpixel $l$ under scale $s$ is computed as:
\begin {equation}\label{eq:mean}
m_{sup_l}^{s}=\frac{1}{P_l}\sum\nolimits_{p=1}^{P_l}\bar{m}_{p}^{s}
\end {equation}
where $P_l$ denotes the number of pixels in superpixel $l$, and $m_{sup_l}^{s}\in\mathbf{M}_{s}^{sup}$. $\bar{m}_{p}^{s}$ is the element in $\bar{\mathbf{M}}_{s}$. The other strategy called ``maxa" is:
\begin {equation}\label{eq:abs}
m_{sup_l}^{s}=\bar{m}_{p_0}^{s}, p_0=\arg\max \limits_{p=1,\cdots, P_l}({\rm abs}(\bar{m}_{p}^{s}-\theta))
\end {equation}
where $\bar{m}_{p}^{s}\in[0,1]$, so we set $\theta=0.5$. With the superpixel-level graph and superpixel-level maps at hands, it is easy to fuse the maps by minimizing the energy function in \cite{korus2016multi}:
\begin {equation}\label{eq:ef}
\frac{1}{S}\sum_{i=1}^{N}\sum_{\{s\}}E_{\tau}(c_i^{(s)},t_i)+\alpha\sum_{i=1}^{N}t_i+\sum_{i=1}^{N}\sum_{j\in\Xi_i}\beta_{ij}|t_i-t_j|
\end {equation}
where $S$ is the number of candidate possibility maps. In our segmentation-based method, $N$ is the number of elements in $\mathbf{M}_{s}^{sup}$, $t_i=1$ denotes tampered units, and $c_i^{(s)}$ denotes the element of the input candidate map with analysis windows of size $s$, i.e. $c_i^{(s)}=m_{sup_l}^{s}$. The three terms can penalize differences of different possibility maps, bias the decision towards the hypotheses and encode a preference towards piecewise-constant solutions. For space limitations, the detailed definitions and discussions of the terms are not provided here, readers can kindly refer to the seminal work \cite{korus2016multi} for details. In terms of the parameters in the energy function, we adopt the default settings of the codes provided by \cite{korus2016multi}.

\section{Experimental evaluation}
\label{sec:Experiments}

Experiments are conducted on two publicly available datasets. In Section \ref{ssec:ExperimentsIFS}, we introduce the experimental results on the image corpus provided in the IFS-TC Image Forensics Challenge (IFS-TC) \cite{IFSTC}. In Section \ref{ssec:ExperimentsRTD}, experiments are conducted on Realistic Tampering Dataset (RTD) \cite{korus2016multi,Korus2016WIFS}.

\subsection{Experiments on IFS-TC}
\label{ssec:ExperimentsIFS}

In the IFS-TC image dataset, there are two sets of images, i.e. $450$ images in the training set with corresponding human-labeled ground truths, and $700$ testing images without ground truths. The scores on the testing set have to be computed by the system provided by the IFS-TC challenge. Thus, in order to test the methods locally, we randomly select $368$ images for training and $75$ images for testing ($7$ images are deserted for imperfect ground truths) from the training set of IFS-TC. For the sake of clarity, the image set of $368$ images is called \emph{sub-training set}, the image set of $75$ images is called \emph{testing set-1}, and the testing set of IFS-TC with $700$ images is called \emph{testing set-2}.

During the patches generation, we also adopt the sliding window manner. The sliding window with a fixed scale slides across the full image. We set the stride $st$ as $8$ to get plenty of sampled patches. In the training set, the tampered areas are marked as the ground truths, we can sample patches based on whether they contain tampered pixels. In \cite{li2017image}, the patches tampered with $10\%$ to $90\%$ are regarded as fake patches for that discriminative features mostly appear around the contours of manipulated regions, we also adopt this strategy. The rates of the tampered areas in the full images differ greatly. In some images, more than ten thousand patches can be generated, while in some images, no patch can be generated. The imbalance of patches distribution can lead to overfitting, so we set an upper threshold $T$. While more than $T$ patches are generated, we randomly select $T$ patches, and we set $T=500$ to make sure that we sample a similar number of patches on most images. With the sliding window sampling manner, no patch can be generated for some images. For those images, we resample patches which are centered at the tampered areas. If the tampered rates of patches are satisfied, the patches are selected. After the fake patches are generated, we sample the same number of pristine patches in the same images, and the pristine patches do not have any tampered pixels. With $5$ groups of sampled patches of scales as $\{32,48,64,96,128\}$, $5$ independent CNNs can be trained, and the CNNs are trained on the \emph{sub-training set}.

Our method is implemented via Caffe and Matlab. Minibatch gradient descent is adopted for training, the momentum is $0.99$ and weight decay is $0.0005$. The learning rate is initialized to $0.001$ and scheduled to decrease $10\%$ for every $8000$ iterations. The convolution kernels are initialized by random numbers generated from zero-mean Gaussian distribution with standard deviation of $0.01$, and bias learning is disabled. The parameters in the fully-connected layer are initialized using ``Xavier". Note that the input patches for the CNNs should all subtract the mean values of each channel.

\begin{figure}[t]
  \centering
  \centerline{\includegraphics[width=7.2cm]{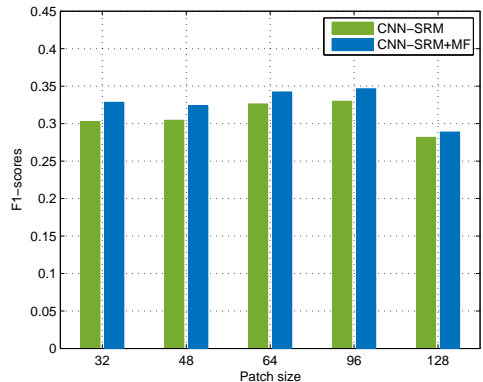}}
\caption{\ The F1-scores of CNNs with input patches of different sizes on IFS-TC \emph{testing set-1}.
}\label{fig:scalesmfval}
\end{figure}

We summarize localization performance as an average F1-score \cite{li2017image}. As shown in Table \ref{table:SCRMCNN}, the comparisons between SCRM+LDA (codes provided by \cite{goljan2014rich,kodovsky2012ensemble}) and different variants of CNNs are conducted. ``MF" denotes mean filtering, and it can certainly improve the F1-scores based on the experimental observation. The results in Figure \ref{fig:scalesmfval} also corroborate that, and the main reason of the improvement achieved by MF smoothing is that the map for an image tends to be smoother without mosaic artifacts caused by sliding-window operations. The training procedure of SCRM+LDA takes too much time. Although we have a powerful CPU, it takes almost $4$ days. Furthermore, its average computing time on the images is also unacceptable. With the same patch size (64) and stride (16), the computing time of CNN is $1/340$ of SCRM+LDA. CNN-SRM denotes the CNN with fixed SRM base filters, CNN-C-SRM denotes constrained filters with SRM initialization and the base filters of CNN-C-GAU are constrained filters with Gaussian initialization. It can be seen that CNN-SRM can achieve higher F1-scores. Because there are many zero values in the SRM base filters, it is also more efficient than CNN-C-SRM and CNN-C-GAU.

\begin{table}[!t]
\renewcommand{\arraystretch}{1.3}
\caption{The comparisons on the IFS-TC \emph{testing set-1}. Time-1 denotes the training time, and Time-2 denotes the average computing time.}
\label{table:SCRMCNN}
\centering
\scriptsize
\begin{tabular}{c | c c c c c}
\hline
Method & Size & Stride & Time-1 (s) & Time-2 (s) & F1-score \\
\hline
SCRM+LDA & 64 & 16 & \multirow{2}{*}{$3.20 \times 10^5$} & 2854.75 & 0.2847 \\
SCRM+LDA+MF & 64 & 16 & & 2855.05 & 0.3123 \\
\hline
CNN-SRM & 64 & 8 & \multirow{2}{*}{3376.07} & 17.11 & 0.3263 \\
CNN-SRM+MF & 64 & 8 & & 17.32 & \bf{0.3423} \\
\hline
CNN-SRM+MF & 64 & 16 & 3376.07 & 8.38 & 0.3354 \\
CNN-C-SRM+MF & 64 & 8 & 3843.03 & 31.66 & 0.2816 \\
CNN-C-GAU+MF & 64 & 8 & 3849.09 & 31.71 & 0.2718 \\
\hline
\end{tabular}
\end{table}

\begin{figure}[t]
  \centering
  \centerline{\includegraphics[width=7.2cm]{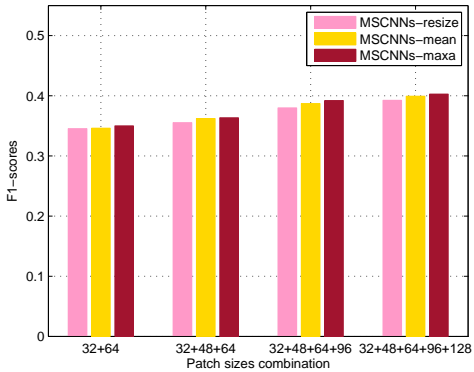}}
\caption{\ The F1-scores of MSCNNs with different combinations on IFS-TC \emph{testing set-1}.
}\label{fig:scalescombval}
\end{figure}

For the good performance of CNN-SRM, we adopt this form of CNN for multi-scale analyses, and the stride is set as $8$. As shown in Figure \ref{fig:scalesmfval}, CNNs with scales as $64$ and $96$ can achieve higher scores, and no single-scale CNN can achieve a score higher than $0.35$. Nevertheless, as shown in Figure \ref{fig:scalescombval}, the multi-scale analysis can improve the performance significantly. As an alternative, we resize the maps, and conduct maps fusion on the resized pixel-level maps directly. Let $w$ and $h$ denote the width and height of the maps, if $w>2000$ and $h>2000$, the map is reduced to $1/10$ of the original map; if $w<1000$ and $h<1000$, the map is reduced to $1/2$; otherwise, it is reduced to $1/4$. We call this kind of method as ``MSCNN-resize".  ``mean" and ``maxa" represent the two strategies for superpixel-level tampering possibility maps generation.  It can be seen that MSCNNs can achieve higher scores with more scales, and MSCNNs-maxa can achieve a higher score than MSCNNs-resize and MSCNNs-mean. In MSCNNs-maxa and MSCNNs-mean, all the images are empirically segmented into $4000$ superpixels, and adaptive segmentation strategies for maps fusion need further research in the future.

Subsequently, we adopt five single-scale CNNs and the $5$-scale MSCNNs to test on the \emph{testing set-2}. As shown in Table \ref{table:IFSTCTESET}, the right side presents the results of different variants of our method, and the left side presents results of the state-of-the-art methods for splicing detection. In another word, the compared methods are not designed for some particular cases, e.g. copy-move forgery detection, and can be utilized to detect any splicing forgeries. Their results are borrowed from their papers \cite{li2017image,gaborini2014multi,cozzolino2014image}. SCRM+LDA adopts the sliding window manner with the scale of $64$, and our CNN with $s=64$ can achieve better performance than SCRM+LDA. Multi-scale analyses can greatly improve the performance of CNNs, and MSCNNs-maxa can achieve a similar F1-score as the winner of IFS-TC challenge \cite{cozzolino2014image} ($0.4063$ vs. $0.4072$). The winner makes use of three different clues, while MSCNNs-maxa only utilizes features extracted by CNNs and can be further improved by combining other clues.

\begin{table}[!t]
\renewcommand{\arraystretch}{1.3}
\caption{Results on the IFS-TC \emph{testing set-2}.}
\label{table:IFSTCTESET}
\centering
\scriptsize
\begin{tabular}{c c | c c}
\hline
Method & F1-score & Variant & F1-score \\
\hline
S3+SVM \cite{cozzolino2014image} & 0.1115 & CNN-SRM32MF & 0.3436 \\
S3+LDA \cite{li2017image} & 0.1737 & CNN-SRM48MF & 0.3526 \\
PRNU \cite{gaborini2014multi} & 0.2535 & CNN-SRM64MF & 0.3570 \\
SCRM+LDA \cite{li2017image} & 0.3458 & CNN-SRM96MF & 0.3423 \\
 & & CNN-SRM128MF & 0.3135 \\
 & & MSCNNs-resize & 0.4014 \\
 & & MSCNNs-mean & 0.4025 \\
 & & MSCNNs-maxa & \bf{0.4063} \\
\hline
\end{tabular}
\end{table}

We evaluate the computing time on the \emph{testing set-2} in which the sizes of images vary from $922\times691$ to $4752\times3168$ (most images are around $1024\times768$). Experiments are conducted on a machine with Intel(R) Core(TM) i7-5930K CPU $@$ 3.50GHz, $64$GB RAM and a single GPU (TITAN X). As shown in Table \ref{table:IFSTCTESETTIME}, the computing time of $5$-scales MSCNNs is around $60$ s for most images. The MF and Fusion (including SLIC) procedures are implemented on CPU which can be further accelerated by implementing on GPU.

\begin{table}[!t]
\renewcommand{\arraystretch}{1.3}
\caption{Computing time on IFS-TC \emph{testing set-2}.}
\label{table:IFSTCTESETTIME}
\centering
\scriptsize
\begin{tabular}{c c c c c c c}
\hline
 & & 32 & 48 & 64 & 96 & 128 \\
\hline
 \multirow{2}{*}{CNNs} & Average time (s) & 15.47 & 15.10 & 17.74 & 19.21 & 19.56 \\
 & Median time (s) & 7.96 & 7.67 & 8.89 & 9.33 & 9.20 \\
\hline
 \multirow{2}{*}{MF} & Average time (s) & 0.08 & 0.13 & 0.19 & 0.37 & 0.63 \\
 & Median time (s) & 0.04 & 0.07 & 0.11 & 0.20 & 0.35 \\
\hline
\hline
&    \multicolumn{6}{c}{multi-scales fusion: 32+48+64+96+128} \\
\hline
 \multirow{2}{*}{Fusion} & Average time (s) &   \multicolumn{5}{c}{20.88} \\
 & Median time (s) &   \multicolumn{5}{c}{11.75} \\
\hline
\end{tabular}
\end{table}

\subsection{Experiments on RTD}
\label{ssec:ExperimentsRTD}

The RTD dataset contains $220$ realistic forgeries created by hand and covers various challenging tampering scenarios involving both object insertion and removal. The images were captured by four different cameras: Canon 60D (C60D), Nikon D90 (ND90), Nikon D7000 (ND7000), Sony $\alpha$57 (S57). All images are $1920\times1080$ px RGB uint8 bitmaps stored in the TIFF format \cite{korus2016multi,Korus2016WIFS}. Each kind of camera contains $55$ images, and we randomly select $27$ as the \emph{training set}, and the left $28$ images compose the \emph{testing set}. In another words, there are $108$ images in the \emph{training set} and $112$ images in the \emph{testing set}. We adopt the same manner to sample patches on RTD, readers can refer to Section \ref{ssec:ExperimentsIFS} for details.

\begin{figure}[t]
  \centering
  \centerline{\includegraphics[width=7.2cm]{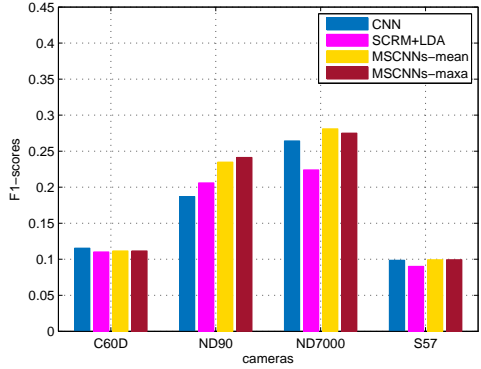}}
\caption{\ The F1-scores on RTD \emph{testing set}. All the models are trained on the \emph{sub-training set} of IFS-TC.
}\label{fig:ifsrtdtest}
\end{figure}

Firstly, we adopt the models trained on the \emph{sub-training set} of IFS-TC to test on the RTD \emph{testing set}. The CNN is the model based on CNN-SRM and mean filtering, and the results of SCRM+LDA are also processed by mean filtering. The size of the sliding window is $64\times64$, and the stride is set as $16$ for fair comparison. The models based on MSCNNs are the $5$-scale models as above mentioned. As shown in Figure \ref{fig:ifsrtdtest}, the performance of all the models decline than the performance on IFS-TC. It proves that both CNN and SCRM+LDA tend to be sensitive to the training sets for that the images may be captured from different cameras and the quality of manipulations may be different. In a different dataset, MSCNNs can still achieve better performance.

\begin{figure}[t]
  \centering
  \centerline{\includegraphics[width=7.2cm]{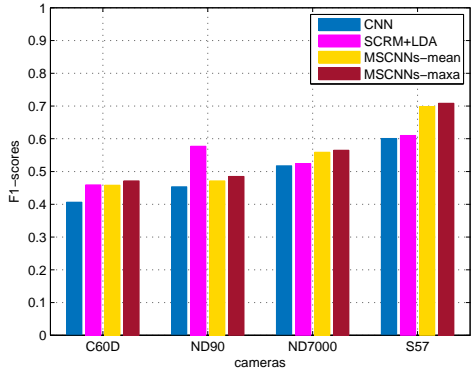}}
\caption{\ The F1-scores on RTD \emph{testing set}. All the models are trained on the \emph{training set} of RTD.
}\label{fig:rtdrtdtest}
\end{figure}

Then, models trained on the RTD \emph{training set} are compared. As shown in Figure \ref{fig:rtdrtdtest}, it can be seen that the performance of CNN is worse than SCRM+LDA. However, with the help of multi-scale analyses, MSCNNs can achieve better performance than SCRM+LDA except for results on ND90. Furthermore, the CNN and MSCNNs are very efficient, the average computing time of CNN is $6.58$ s, and the computing time of MSCNNs is $34.62$ s ($5$ CNNs on GPU) $+30.36$ s (the fusion procedure on CPU), while SCRM+LDA takes $2220.45$ s per image. Thus, MSCNNs is a better alternative of SCRM+LDA in the image forgery localization tasks.

\section{Conclusions}
\label{sec:DiscussionConclusion}

In this paper, a novel forgery localization method based on Multi-Scale Convolutional Neural Networks is proposed. CNNs for color patches of different scales are well designed and trained as a set of forgery detectors. Then, segmentation-based multi-scale analysis is utilized to dig out the information given by the different-scale analyses. Full experiments on the publicly available datasets demonstrate the effectiveness and efficiency of the proposed method named MSCNNs. Although the proposed method can achieve the state-of-the-art performance, it still has a long way to go for real applications. The robustness of existing works against post compression, manipulation qualities and camera models still needs to be further studied. In the future, MSCNNs can also be improved by adopting more powerful CNNs.

\bibliographystyle{IEEEbib}
\bibliography{mybibfile}

\end{document}